\documentclass{article}
\usepackage{iclr2024_conference} % Required for inserting images
\usepackage{graphicx}
\usepackage{subfigure}
\usepackage{booktabs}
\usepackage{tabularx}
\usepackage{geometry}
\usepackage{longtable}
\usepackage{changepage}
\usepackage{hyperref}
\usepackage{amsmath}

\title{Semi-Supervised Domain Adaptation for Wildfire Detection\centering}
\author{
    JooYoung Jang$^{1,2}$ \and 
    Youngseo Cha$^{1}$ \and 
    Jisu Kim$^{1}$ \and 
    SooHyung Lee$^{1}$ \and
    Geonu Lee$^{1}$ \and
    Minkook Cho$^{1}$ \and 
    Young Hwang$^{1}$ \and 
    Nojun Kwak$^{2*}$ \and
    \smallskip
    $^1$Alchera, South Korea~~~
    $^2$Seoul National University, South Korea \\
    \tt\small jyjang1090@snu.ac.kr, \{ys.cha, js.kim, \_shlee, gu.lee\}alcherainc.com, nojunk@snu.ac.kr
}

\date{January 2024}

\begin{document}
\maketitle
\begin{abstract}
% Recently not only the frequency but also the intensity of wildfires have \nj{increased} worldwide mainly due to climate change \citep{sath2023wildfire}. In this paper, we propose a \nj{new protocol for wildfire detection,} semi-supervised Domain Adaptation for object detection, \nj{and the corresponding dataset} that could be used for academic\nj{s} and industries. Our dataset covers 30x diverse labeled scenes using \nj{the} current largest benchmark wildfire dataset, HPWREN \cite{hpwren}, and also suggest\nj{s} new labeling policy for detecting wildfire. Motivated by \cite{liu2018coordconv}, we suggest a strong baseline, Location-Aware object detection for semi-supervised Domain Adaptation (LADA) using \nj{a} teacher-student \cite{liu2021unbiased} based self-supervised Domain Adaptation framework \nj{that is capable of} extracting translational variance features of wildfire. Our framework outperforms \nj{the} baseline by an absolute margin of 3.8\nj{\%} mean Average Precision (mAP) in \nj{the} HPWREN wildfire dataset.
% ChaptGPT-4v
Recently, both the frequency and intensity of wildfires have increased worldwide, primarily due to climate change \citep{sath2023wildfire}. In this paper, we propose a novel protocol for wildfire detection, leveraging semi-supervised Domain Adaptation for object detection, accompanied by a corresponding dataset designed for use by both academics and industries. Our dataset encompasses 30 times more diverse labeled scenes for the current largest benchmark wildfire dataset, \cite{hpwren}, and introduces a new labeling policy for wildfire detection. Inspired by \cite{liu2018coordconv}, we propose a robust baseline, Location-Aware Object Detection for Semi-Supervised Domain Adaptation (LADA), utilizing a teacher-student \citep{liu2021unbiased} based framework capable of extracting translational variance features characteristic of wildfires. With only using 1\% target domain labeled data, our framework significantly outperforms our source-only baseline by a notable margin of 3.8\% in mean Average Precision on the HPWREN wildfire dataset. Our dataset is available at \href{https://github.com/BloomBerry/LADA}{https://github.com/BloomBerry/LADA}.
\end{abstract}

\section{Introduction}
Wildfires contribute to and are exacerbated by global warming, leading to significant economic losses and ecological damage \citep{sath2023wildfire, Lindsey2023wildfire}. Such impacts can be mitigated through the early detection of wildfires, enabling firefighters to intervene promptly. For effective mitigation, wildfire detection systems must achieve high accuracy and maintain low false positive rates \citep{Ranadive2022Firescout}. 

\begin{figure}
\centering
\includegraphics[width=0.75\linewidth]{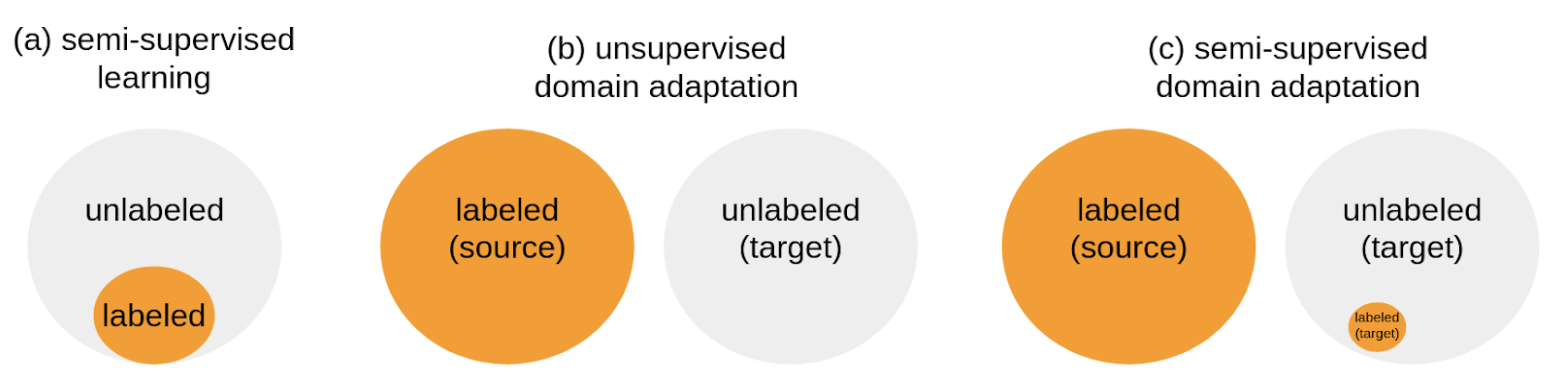}
\caption{semi-supervised Domain Adaptation}
\label{fig:DA}
\end{figure}

However, applying fire detection in real-world scenarios present challenges, including a domain shift between the training and testing environments that can degrade detection performance \citep{Yoo2022OADA}. Even worse, acquiring a large volume of labeled data for the target domain is particularly challenging for infrequent events like wildfires \citep{Kim2022MUM}. To address these challenges, this paper introduces a new protocol, semi-supervised domain adaptation(SSDA) for wildfire detection. To the best of our knowledge, this is the first paper to apply SSDA for object detection task. As depicted in Fig.\ref{fig:DA}, SSDA is a combination of semi-supervised learning (SSL) and unsupervsied domain adaptation (UDA) task. It uses large amount of source domain images, while uses minimal set of target labeled images alongside a substantial corpus of target unlabeled images. SSDA setting is practical for real-world application considering labeling cost and performance. \citep{Yu2023SLA}

\noindent\rule{\textwidth}{1pt} % Creates a horizontal line that spans the text width
$^{*}$corresponding author: Nojun Kwak (nojunk@snu.ac.kr) % Italicized text below the line

This work makes two primary contributions. First, we introduce new labels for wildfire detection tasks, increasing label diversity by thirtyfold compared to existing labels within the HPWren dataset, as in Table \ref{tab:label_policy}. We classified source domain as previous publicily available labels and target domain as new labeled set we suggested in this paper. Second, we present a novel approach to learn translational variance characteristics of wildfires called Location-Aware Semi-Supervised Domain Adaptation (LADA) framework, which integrates Coordinate Convolution \citep{liu2018coordconv} with a scale-aware Faster R-CNN \citep{Chen2021SAFRCNN}. Our result demostrate inhanced performance accross various SSDA protocols from current state-of-the-art UDA framework \citep{Hoyer2022MIC}.

\textbf{Related work.} SSDA approach seeks to reduce the domain-gap between source and target using consistency regularization loss \citep{Yu2023SLA}, or use cutmix augmentation for domain-mixing loss \citep{Chen2021DDM}. However, those methods neither applied for object detection task. Our method uses consistency regularization loss using masked augmentation similar to \citep{Hoyer2022MIC}.

\section{Methods}
\subsection{Proposed Dataset}
\begin{figure}[ht]
    \centering
    \begin{subfigure}{}
        \centering
        \includegraphics[width=0.3\linewidth]{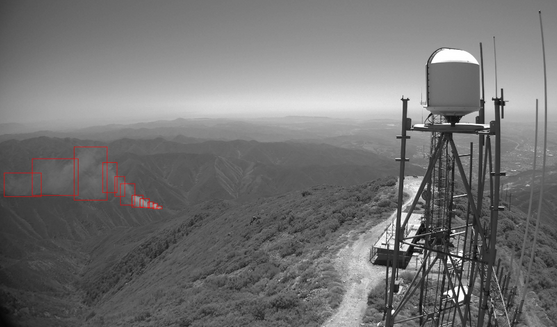}
        \label{fig:subfig_a}
    \end{subfigure}
    \begin{subfigure}{}
        \centering
        \includegraphics[width=0.3\linewidth]{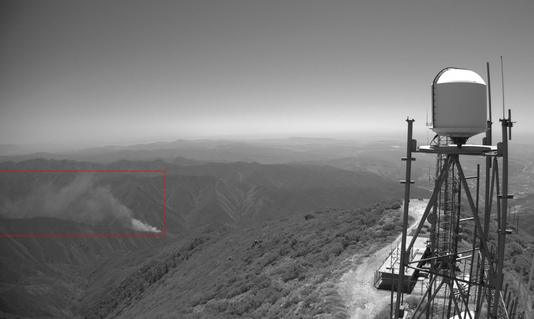}
        \label{fig:subfig_b}
    \end{subfigure}

    \caption{Original HPWREN labeled image (Left) vs. Proposed labeled image (Right)\centering}
    \label{fig:Labeling policy comparison: previous vs. purposed}
\end{figure}

% In this paper, we propose labels for HPWREN dataset\cite{hpwren}. According to \cite{hpwren2}, this dataset suggest wildfire benchmark for individual researchers. However, there are two main obstacles to apply this benchmark directly for research. First, the quantity and quality of labels are not diverge, which raise overfitting issues. To be specific, it only suggests 609 labeled images with 9 scenes. Secondly, it labels smokes as a splitted bounding boxes, which requires a lot of time to label. However, we find that simply merging the bounding boxes enhances not only simplicity of labeling but detection performance as show in Fig. \ref{fig:Labeling policy comparison: previous vs. purposed}. The results are in sec. \ref{sec:results}. \\
% By overcoming previously mentioned problems, we propose a new benchmark of semi-supervised Domain Adaptation for object detection in Table \ref{tab:hpwren}. Inspired by \cite{Li2024GABC}, we suggest three protocols, namely 0.5\%, 1.0\%, and 3.0\% meaning target domain labeled ratio among total target data. In this benchmark, source domain consists of 9 sub-directories which the labels exist in HPWREN \cite{hpwren2} homepage, and 274 sub-directories remaining as target domain. In this way, source and target domain has domain shift since they don't include common environment. More information of our dataset is covered in Appendix \ref{app:dataset}.
% ChaptGPT-4v
In this paper, we propose a refined set of labels for the \citet{hpwren} dataset. This dataset serves as a benchmark for wildfire detection tailored to individual researchers. However, direct application of this benchmark for research encounters two primary obstacles. First, the diversity and quality of the labels are limited, leading to potential overfitting issues. Specifically, the dataset comprises only 609 labeled images across 9 scenes. Second, the practice of labeling smoke with separated bounding boxes demands considerable time and efforts for annotation. We have discovered that merging these bounding boxes not only simplifies the labeling process but also improves detection performance, as illustrated in Fig. \ref{fig:Labeling policy comparison: previous vs. purposed}. Detailed results are presented in Section \ref{sec:results}.

To address these challenges, we introduce a new benchmark for semi-supervised domain adaptation in object detection, detailed in Table \ref{tab:hpwren}. Inspired by \cite{Li2024GABC}, we propose three protocols, 0.5\%, 1.0\%, and 3.0\%, representing the ratio of labeled data in the target domain relative to the total dataset. In this benchmark, the source domain comprises 9 sub-directories with labels available on the \cite{hpwren} homepage, while 274 sub-directories are designated as the target domain. This configuration results in a domain shift, as the source and target domains do not share a common environment. 
% Further details about our dataset are provided in Appendix \ref{app:dataset}.

% table
\begin{table}[h]
  \centering
  \caption{Number of sub-directories, and labels for HPWREN dataset}
  \begin{tabular}{|c|c|c|c|}
    \hline
      & \textbf{Previous labels} & \textbf{Proposed labels} & \textbf{Total HPWREN} \\
    \hline
    \textbf{Number of sub-directories} & 9 & 283 & 342 \\
    \hline
    \textbf{Number of images} & 609 & 2,575 & 27,174 \\
    \hline
    
  \end{tabular}
  \label{tab:hpwren}
\end{table}

% table 3
\begin{table}[h]
  \caption{semi supervised domain adaptation for wildfire detection protocol}
  \begin{tabular}{|c|c|c|c|c|c|}
    \hline
      & \textbf{source} & \textbf{target 0.5\%} & \textbf{target 1.0\%} & \textbf{target 3.0\%} & \textbf{target val}\\
    \hline
    \textbf{forground images} & 309 & 44 & 94 & 257 & 451 \\
    \hline
    \textbf{background images} & 300 & 58 & 111 & 359 & 630\\
    \hline
    \textbf{total images} & 609 & 102 & 205 & 616 & 1,081\\
    \hline
  \end{tabular}
  \label{tab:ssda}
\end{table}

% figure 

\subsection{Location-Aware semi-supervised Domain Adaptation}

% ChatGPT-v4
\textbf{Preliminary.} In this study, we tackle the challenge of early wildfire detection by leveraging object detection frameworks \citep{Chen2021SAFRCNN}. Image samples are denoted by $\mathbf{x}_s=(x_i)^{N_s}_{i=1}, \mathbf{x}_{tl}=(x_i)^{N_{tl}}_{i=1}$ and corresponding bounding-box labels $\mathbf{y}_s=(y_i)^{N_s}_{i=1}, \mathbf{y}_{tl}=(y_i)^{N_{tl}}_{i=1}$ are utilized as input to the model, where $N_s, N_{tl}$ is the number of labeled samples for source and target domain each. The label consists of class and bounding boxes $y_i=(c, x, y, w, h)$ where $c, x, y, w$ represents class index, center points, width and height of the box. In addition, the pseudo bounding-box labels $\mathbf{u}=(u_i)^{N_{tu}}_{i=1}$ are constructed when the confidence score is bigger than the upper threshold $p_u > \tau_u$, or smaller than the lower threshold $p_u < \tau_l$ where $\tau_u, \tau_l$, and $N_{tu}$ represents upper, lower confidence threshold, and number of unlabeled target samples, each.

\textbf{Pseudo Labeling.}  We utilize large amount of unlabeled target data with a teacher-student paradigm \citep{liu2021unbiased} augmented with Masked Image Consistency (MIC) Loss \citep{Hoyer2022MIC}. Built upon that, we changed pseudo label filter to use reliable background images implemented by very low probability score in order to train backbround images, as shown in equation \ref{pseudo_labeling}. $\hat{y}_i$ reprsents \textit{i}$_{th}$ unlabeled target sample index. We used $\tau_u=0.8, \tau_l=0.05$ for all of our experiments. We find that it is helpful especially for 0.5\%, 1.0\% SSDA protocols which lacks of highly reliable positive images. Further information is in Appendix \ref{app:background}. 

\begin{figure}[ht]
    \centering
    \begin{subfigure}{}
        \centering
        \includegraphics[width=0.40\linewidth]{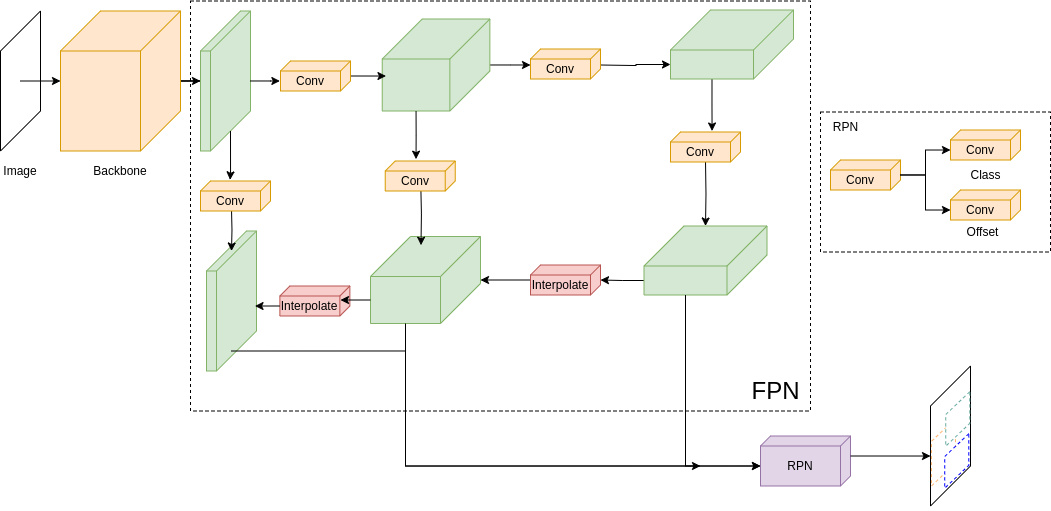}
        \label{fig:subfig_a}
    \end{subfigure}
    \begin{subfigure}{}
        \centering
        \includegraphics[width=0.45\linewidth]{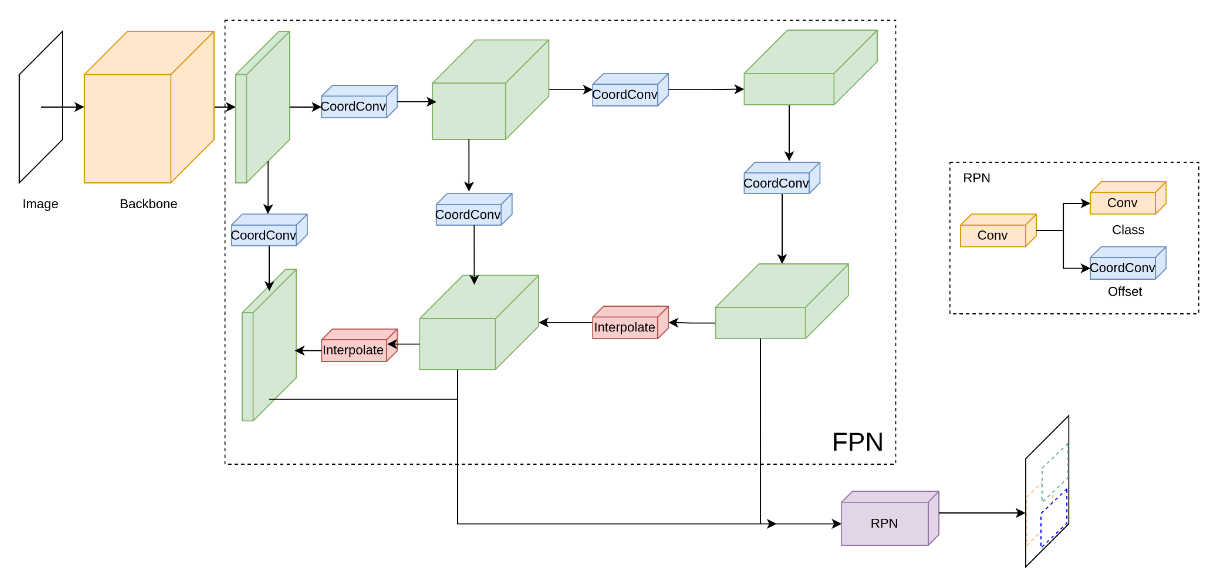}
        \label{fig:subfig_b}
    \end{subfigure}
    \caption{Location Aware semi-supervised Domain Adaptation Network. We omitted the second stage regression and classification heads for simplicity. \centering\centering}
    \label{fig:LADA}
\end{figure}

% Upon this baseline, we added Coordinate Convolution \cite{liu2018coordconv} as shown in Fig. \ref{fig:LADA}. We used coordinate convolution layer in the Feature Pyramid Network (FPN) \cite{Lin2017FPN} and Region Proposal Network (RPN) \cite{Ren2015FRCNN} because wildfire has location-dependent features. For example, wildfire is usually corn shape, enlaring its size upward. Also, it doesn't break out in upper side of the image which is mainly covered by sky. By simply adding coordinate information with two channels, x and y, this allows the model to catch such translational variance characteristics with negligible computational cost.
\textbf{Translation Variance Features.} Wildfires typically don't occur in speacific area, such as skies or lakes, or it shows a location-dependent shapes. For instance, they seldom occur in the upper portion of the images, which are predominantly occupied by the sky, while many of them has conical shape, expanding vertically. In order to utilize such characteristics, the Coordinate Convolution layer \citep{liu2018coordconv} was incorporated into both the Feature Pyramid Network (FPN) \citep{Lin2017FPN} and the Region Proposal Network (RPN) \citep{Ren2015FRCNN}, as illustrated in Fig. \ref{fig:LADA}. Coordinate convolution layer embed coordinate information through two channels, \(x\) and \(y\) to the original convolution layers, and the new added channels enable the model to capture such translational variance features at a minimal computational cost. We didn't add Coordinate convolution layer into the backbone as suggested in the \citep{liu2018coordconv}, since it did not show good performance.

\begin{figure}
\includegraphics[width=0.8\linewidth]{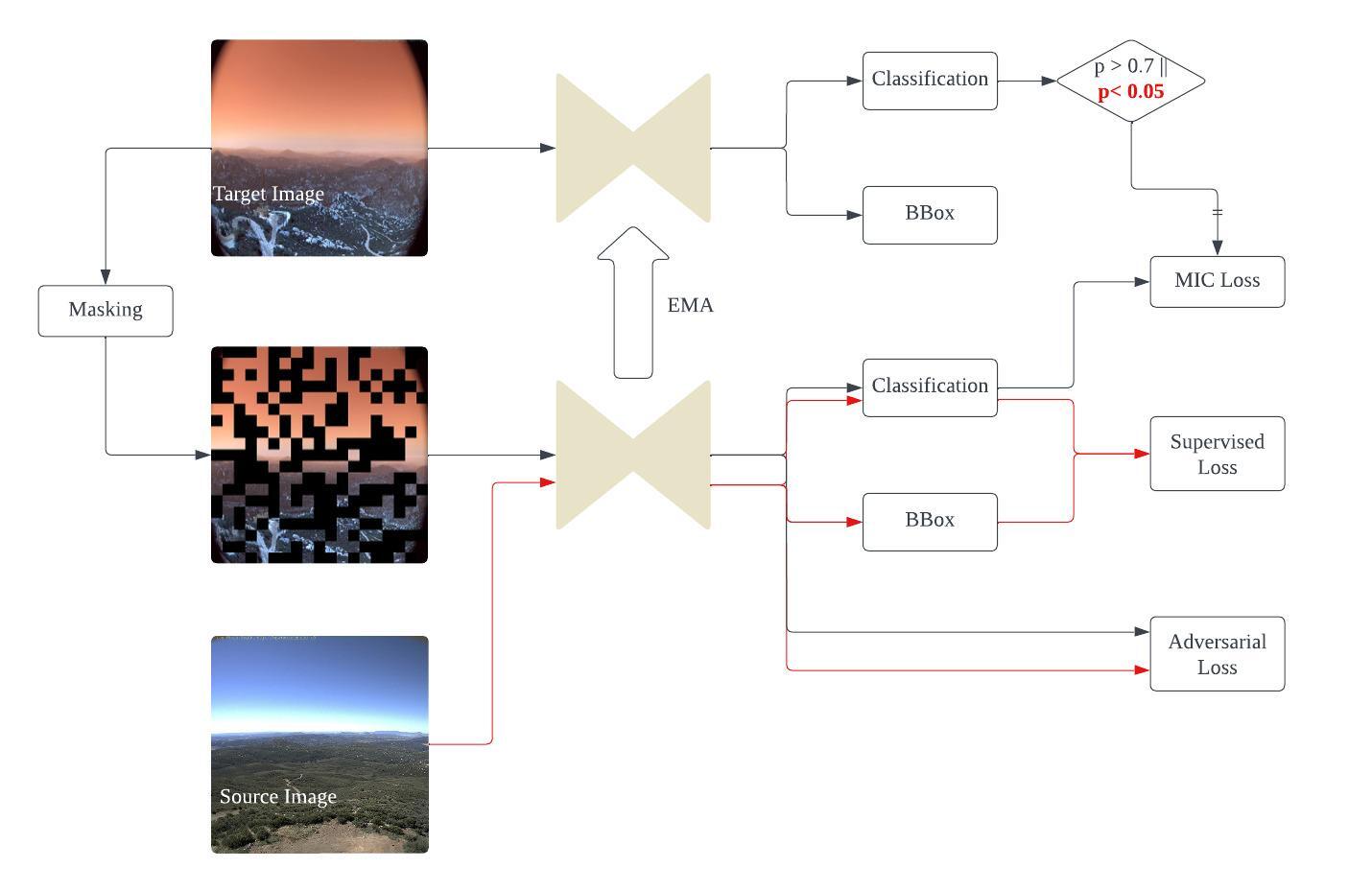}
\centering
\caption{Overall Diagram of our training process. We also use background images for training.\centering}
\label{fig:overall}
\end{figure}

\textbf{Location Aware Domain Adaptation.} The comprehensive diagram of our training process is depicted in Fig. \ref{fig:overall}. The student model is trained using both supervised and unsupervised losses, whereas the teacher model is periodically updated through an Exponential Moving Average. Unsupervised losses consist of masked image consistency loss, pseudo label based Cross entropy loss, and adversarial loss. The former allows the model to learn consistent output from masked image to original image, which allows the model to learn robust predictions even in such randomly masked images. Pseudo labeling loss, on the other hand, make use of a large amount of unlabeled target images to train as supervised learning. Finally, adversarial loss aligns the source and target domain features in the backbone in order to reduce domain gap in three levels. More training detail information is available in Appendix \ref{app:Training details}.

% figure 3

% figure 2

% equation 1
\begin{equation}
\left\{
\begin{aligned}
& \text{$\hat{p_i} > \tau_u$} \\
& \text{$\hat{p_i} < \tau_l$}
\end{aligned}
\right.
\label{pseudo_labeling}
\end{equation}

%%%%%%%%%%%%%%%%%%%%
\section{Results}
\label{sec:results}
% We trained our SSDA model in 2 stage. In first stage, we only used source data. In second stage, we used both source labeled data, target labeled, and unlabeled data.\\
% We first compared the original HPWREN labels against proposed labels. As shown in Table \ref{tab:label_policy}, our merged bounding box label shows 6.4 to 10.9 mAP improvement. Therefore, we suggest to use our merged bounding box labels. \\
%ChatGPT-v4
Our model is trained in two stages. In the first stage, the training exclusively utilized source data. In the subsequent stage, a combination of labeled source data, labeled target data, and unlabeled data was employed. We conducted an initial comparison between the original HPWREN labels and our proposed labeling approach. As detailed in Table \ref{tab:label_policy}, our approach, which utilizes merged bounding boxes, improves up to 10.9 mean Average Precision@0.5:0.95 (mAP) over the original labels. Based on these results, we advocate for the adoption of our merged bounding box labeling strategy in wildfire detection.

% table 2
\begin{table}[h]
  \centering
  \caption{Comparison between original \& proposed labeling policy at mAP/mAP@0.5\centering}
  \begin{tabular}{|c|c|c|c|}
    \hline
      & \textbf{0.5\%} & \textbf{1.0\%} & \textbf{3.0\%} \\
    \hline
    \textbf{original label} & 1.5/7.0 & 2.8/12.9 & 7.9/29.6 \\
    \hline
    \textbf{merged label} & 7.9/24.0 & 10.2/31.5 & 18.8/48.4 \\
    \hline
    
  \end{tabular}
  \label{tab:label_policy}
\end{table}

% As shown in Table \ref{tab:source_only}, our model outperforms \cite{Chen2021SAFRCNN} in source-only protocol which only uses source dataset for training while vaildate in target validation dataset.
As presented in Table \ref{tab:results}, our model surpasses the performance of the model proposed by \cite{Chen2021SAFRCNN} in the source-only protocol. This protocol exclusively utilizes the source dataset for training and subsequently validates the model on the target validation dataset. Our model also outperforms \cite{Hoyer2022MIC} in semi-supervised Domain Adaptation protocols. The results indicate that our proposed method catches the translational variance features of wildfire well, leading to better generalization performance.
% More ablation studies about our experiments are in Appendix \ref{app:unlabel_ratio}.

% table 4

\begin{table}[ht]
\centering
\begin{tabular}{|c|l|c|c|c|}
\hline
\textbf{Type} & \textbf{Methods} & \multicolumn{3}{c|}{\textbf{Labeled target images}} \\
 & & \textbf{0.5\%} & \textbf{1.0\%} & \textbf{3.0\%}   \\ % Add & \textbf{1000} if needed
\hline
Source-only & SADA \citep{Chen2021SAFRCNN} & 6.9/21.9 & 9.7/28.7 & 17.8/48.0   \\ % Add & - if needed
 & LADA(ours) & \textbf{7.9/24.0} & \textbf{10.2/31.5} & \textbf{18.8/48.4}   \\
\hline
SSDA & SADA \citep{Hoyer2022MIC} & 9.7/27.3 & 12.3/34.9 & 20.4/53.0   \\ % Add & - if needed
 & LADA(ours) & \textbf{10.0/29.1} & \textbf{14.0/38.0} & \textbf{20.9/52.3} \\
 \hline

\end{tabular}
\caption{Comparison of source-only and SSDA results. (mAP/mAP@0.5)}
\label{tab:results}
\end{table}

% \begin{table}[h]
%   \centering
%   \caption{LADA vs. baseline for source-only protocol}
%   \begin{tabular}{|c|c|c|c|}
%     \hline
%       & \textbf{ssda 0.5\%} & \textbf{ssda 1.0\%} & \textbf{ssda 3.0\%} \\
%     \hline
%     \textbf{SADA} & 6.9 & 9.7 & 17.8 \\
%     \hline
%     \textbf{LADA(ours)} & 7.9 & 10.2 & 18.8 \\
%     \hline
    
%   \end{tabular}
%   \label{tab:source_only}
% \end{table}

% table 5
% \begin{table}[h]
%   \centering
%   \caption{LADA vs. baseline for ssda protocol}
%   \begin{tabular}{|c|c|c|c|}
%     \hline
%       & \textbf{ssda 0.5\%} & \textbf{ssda 1.0\%} & \textbf{ssda 3.0\%} \\
%     \hline
%     \textbf{SADA} & 9.7 & 12.3 & 20.4 \\
%     \hline
%     \textbf{LADA(ours)} & \textbf{10.0} & \textbf{14.0} & \textbf{20.9} \\
%     \hline
    
%   \end{tabular}
%   \label{tab:ssda_result}
% \end{table}
%%%%%%%%%%%%%%%%%%%%

\section{CONCLUSION}
% In this paper, we proposed a new benchmark using semi-supervised Domain Adaptation for object detection which can be used for academic, and industries. The diversity of our labels cover 30x larger than previous wildfire benchmark, and also suggest new labeling policy for wildfire detection. We also suggest strong baseline for our benchmark, named LADA (Location-Aware semi-supervised Domain Adaptation), which characterizes with translational variance features suited for wildfire detection. We hope our research would provide valuable insights for future researchers to tackle semi supervised domain adaptation into wildfire detection.
% ChatGPT-v4
In this paper, we propose a novel benchmark utilizing semi-supervised domain adaptation for object detection, designed to benefit both academia and industry. Our labeling approach introduces a diversity that is thirtyfold greater than that of existing wildfire benchmarks and presents a new labeling policy tailored for wildfire detection. Furthermore, we establish a robust baseline for this benchmark, named LADA (Location-Aware Semi-Supervised Domain Adaptation), distinguished by its capability to capture translational variance features pertinent to wildfire detection.

\section{ACKNOWLEDGEMENTS}
This work was supported by NRF grant (2022R1A5A7026673) funded by MSIT, Korean Government. 
\newpage
% %%%%%%%%% REFERENCES
{\small
\bibliographystyle{iclr2024_conference}
\bibliography{egbib}
}
%% Appendix
\newpage
\begin{appendix}
\section{Dataset}
\label{app:dataset}
This section will give more information of HPWREN dataset, and how source and target domain has been composed. Each image files are named as equation \ref{naming_convention}.

% equation 3
\begin{equation}
YYYYMMDD\_fireName\_cameraName
\label{naming_convention}
\end{equation}

We defined domain shift based on equation \ref{naming_convention}, and simply splitted train and validation set. We splitted target validation set and target train set with 5\%, 95\% by random sampling. We also random sampled 0.5\%, 1.0\%, 3.0\% target labeled dataset among 95\% of target train dataset for each semi-supervised domain adaptaion protocols. Sub-directories for source domain and target domain are summarized in Table \ref{source domain scenes and number of images} to \ref{tab:target_domain11}. However, we noticed that users could  also split based on customized domain shift scenario. For example, we illustrate defining domains with $\textbf{cameraName}$ in Equation \ref{naming_convention}. It is summarized in Table \ref{source domain scenes and number of images using cameraName} to \ref{tab:target domain scenes and number of images for cameraName2}.

\begin{table}[htbp]
\centering
% Define new column types with centered text
\newcolumntype{Y}{>{\centering\arraybackslash}X}
\begin{tabularx}{\textwidth}{>{\hsize=1.5\hsize}Y>{\hsize=0.5\hsize}Y>{\hsize=1.5\hsize}Y>{\hsize=0.5\hsize}Y}
\toprule
\textbf{Scene Description} & \textbf{\small\# of Imgs} & \textbf{Scene Description} & \textbf{\small\# of Imgs} \\
\midrule
20160604\_FIRE\_rm-n-mobo-c & 81 & {\small 20160604\_FIRE\_smer-tcs3-mobo-c} & 81 \\
20160619\_FIRE\_lp-e-iqeye & 41 & 20160619\_FIRE\_om-e-mobo-c & 81 \\
20160619\_FIRE\_pi-s-mobo-c  & 81 & 20160711\_FIRE\_ml-n-mobo-c  & 81 \\
20160718\_FIRE\_lp-n-iqeye & 41 & 20160718\_FIRE\_mg-s-iqeye & 41 \\
20160718\_FIRE\_mw-e-mobo-c & 81 & &  \\
\bottomrule
\end{tabularx}
\caption{Scenes composed for source domain dataset by equation \ref{naming_convention}}
\label{source domain scenes and number of images}
\end{table}

\begin{table}[htbp]
\centering
\begin{tabular}{llll}
\toprule
\textbf{Scene Description} & \textbf{\# of Imgs} & \textbf{Scene Description} & \textbf{\# of Imgs} \\
\midrule
20160722\_FIRE\_mg-s-iqeye & 41 & 20170708\_Whittier\_syp-n-mobo-c & 81 \\
20160722\_FIRE\_mw-e-mobo-c & 81 & 20170708\_Whittier\_syp-n-mobo-m & 80 \\
20161113\_FIRE\_bl-n-mobo-c & 81 & 20170711\_FIRE\_bl-e-mobo-c & 81 \\
20161113\_FIRE\_bm-n-mobo-c & 81 & 20170711\_FIRE\_bl-s-mobo-c & 81 \\
20161113\_FIRE\_bm-w-mobo-c & 81 & 20170711\_FIRE\_bm-s-mobo-c & 64 \\
20170519\_FIRE\_rm-w-mobo-c & 81 & 20170711\_FIRE\_sdsc-e-mobo-c & 81 \\
20170520\_FIRE\_lp-s-iqeye & 81 & 20170711\_FIRE\_sm-n-mobo-c & 81 \\
20170520\_FIRE\_om-s-mobo-c & 55 & 20170713\_FIRE\_smer-tcs8-mobo-c & 77 \\
20170520\_FIRE\_pi-s-mobo-c & 81 & 20170722\_FIRE\_bm-n-mobo-c & 81 \\
20170520\_FIRE\_pi-w-mobo-c & 81 & 20170722\_FIRE\_hp-e-mobo-c & 81 \\
20170609\_FIRE\_sm-n-mobo-c & 81 & 20170722\_FIRE\_mg-n-iqeye & 81 \\
20170613\_FIRE\_bh-w-mobo-c & 81 & 20170722\_FIRE\_so-s-mobo-c & 81 \\
20170613\_FIRE\_hp-n-mobo-c & 81 & 20170807\_FIRE\_bh-n-mobo-c & 78 \\
20170625\_BBM\_bm-n-mobo & 81 & 20170821\_FIRE\_lo-s-mobo-c & 81 \\
20170625\_FIRE\_mg-s-iqeye & 81 & 20170826\_FIRE\_tp-s-mobo-c & 81 \\
\bottomrule
\end{tabular}
\caption{Scenes composed for target domain dataset by equation \ref{naming_convention} (Part. 1)}
\label{tab:target_domain1}
\end{table}

\begin{table}[htbp]
\centering
\begin{tabular}{llll}
\toprule
\textbf{Scene Description} & \textbf{\# of Imgs} & \textbf{Scene Description} & \textbf{\# of Imgs} \\
\midrule
20170901\_FIRE\_om-s-mobo-c & 81 & 20171017\_FIRE\_smer-tcs3-mobo-c & 78 \\
20170927\_FIRE\_smer-tcs9-mobo-c & 81 & 20171021\_FIRE\_pi-e-mobo-c & 81 \\
20171010\_FIRE\_hp-n-mobo-c & 81 & 20171026\_FIRE\_rm-n-mobo-c & 81 \\
20171010\_FIRE\_hp-w-mobo-c & 81 & 20171026\_FIRE\_smer-tcs8-mobo-c & 81 \\
20171010\_FIRE\_rm-e-mobo-c & 81 & 20171207\_FIRE\_bh-n-mobo-c & 81 \\
20171016\_FIRE\_sdsc-e-mobo-c & 81 & 20171207\_FIRE\_bh-w-mobo-c & 77 \\
20171017\_FIRE\_smer-tcs3-mobo-c & 78 & 20171207\_FIRE\_smer-tcs8-mobo-c & 81 \\
20171021\_FIRE\_pi-e-mobo-c & 81 & 20171207\_Lilac\_rm-s-mobo & 81 \\
20171026\_FIRE\_rm-n-mobo-c & 81 & 20180504\_FIRE\_bh-n-mobo-c & 81 \\
20171026\_FIRE\_smer-tcs8-mobo-c & 81 & 20180504\_FIRE\_rm-n-mobo-c & 81 \\
20171207\_FIRE\_bh-n-mobo-c & 81 & 20180504\_FIRE\_smer-tcs10-mobo-c & 81 \\
20171207\_FIRE\_bh-w-mobo-c & 77 & 20180504\_FIRE\_smer-tcs8-mobo-c & 81 \\
20171207\_FIRE\_smer-tcs8-mobo-c & 81 & 20180517\_FIRE\_rm-n-mobo-c & 81 \\
20171207\_Lilac\_rm-s-mobo & 81 & 20180517\_FIRE\_smer-tcs10-mobo-c & 81 \\
20180504\_FIRE\_bh-n-mobo-c & 81 & 20180522\_FIRE\_rm-e-mobo-c & 81 \\
20180504\_FIRE\_rm-n-mobo-c & 81 & 20180602\_Alison\_sp-s-mobo-c & 81 \\
20180504\_FIRE\_smer-tcs10-mobo-c & 81 & 20180602\_Alison\_sp-w-mobo-c & 81 \\
20180504\_FIRE\_smer-tcs8-mobo-c & 81 & 20180602\_FIRE\_rm-n-mobo-c & 81 \\
20180517\_FIRE\_rm-n-mobo-c & 81 & 20180602\_FIRE\_smer-tcs8-mobo-c & 81\\
20180602\_FIRE\_smer-tcs9-mobo-c & 81 & 20180603\_FIRE\_bl-s-mobo-c & 81\\
\bottomrule
\end{tabular}
\caption{Scenes composed for target domain dataset by equation \ref{naming_convention} (Part. 2)}
\label{tab:target_domain2}
\end{table}

\begin{table}[htbp]
\centering
\begin{tabular}{llll}
\toprule
\textbf{Scene Description} & \textbf{\# of Imgs} & \textbf{Scene Description} & \textbf{\# of Imgs} \\
\midrule
20180603\_FIRE\_rm-w-mobo-c & 81 & 20180606\_FIRE\_lo-s-mobo-c & 81 \\
20180603\_FIRE\_smer-tcs8-mobo-c & 81 & 20180606\_FIRE\_ml-s-mobo-c & 81 \\
20180603\_FIRE\_smer-tcs9-mobo-c & 81 & 20180606\_FIRE\_pi-e-mobo-c & 81 \\
20180603\_FIRE\_sm-n-mobo-c & 81 & 20180611\_fallbrook\_rm-w-mobo-c & 81 \\
20180603\_FIRE\_sm-w-mobo-c & 81 & 20180612\_FIRE\_rm-w-mobo-c & 81 \\
20180605\_FIRE\_rm-w-mobo-c & 81 & 20180612\_FIRE\_smer-tcs9-mobo-c & 81 \\
20180605\_FIRE\_smer-tcs9-mobo-c & 81 & 20180614\_Bridle\_hp-n-mobo-c & 81 \\
20180614\_FIRE\_hp-s-mobo-c & 68 & 20180704\_Benton\_hp-n-mobo-c & 81 \\
20180614\_Hope\_wc-e-mobo-c & 81 & 20180706\_FIRE\_sm-e-mobo-c & 81 \\
20180706\_FIRE\_sm-n-mobo-c & 70 & 20180706\_FIRE\_wc-e-mobo-c & 69 \\
20180706\_West\_lp-n-mobo-c & 81 & 20180717\_otay\_om-s-mobo-c & 81 \\
20180718\_FIRE\_syp-w-mobo-c & 81 & 20180719\_Skyline\_sp-n-mobo-c & 81 \\
20180720\_Cinnamon\_wc-e-mobo-c & 81 & 20180720\_FIRE\_syp-w-mobo-c & 81 \\
20180723\_FIRE\_tp-e-mobo-c & 81 & 20180725\_Cranston\_hp-n-mobo-c & 81 \\

\bottomrule
\end{tabular}
\caption{Scenes composed for target domain dataset by equation \ref{naming_convention} (Part. 3)}
\label{tab:target_domain3}
\end{table}

\begin{table}[htbp]
\centering
\begin{tabular}{llll}
\toprule
\textbf{Scene Description} & \textbf{\# of Imgs} & \textbf{Scene Description} & \textbf{\# of Imgs} \\
\midrule
20180725\_Cranston\_sp-e-mobo-c & 81 & 20180806\_FIRE\_mg-s-mobo-c & 78 \\
20180725\_FIRE\_smer-tcs10-mobo-c & 81 & 20180806\_FIRE\_vo-w-mobo-c & 81 \\
20180726\_FIRE\_so-n-mobo-c & 81 & 20180806\_Holy\_sp-s-mobo-c & 72 \\
20180726\_FIRE\_so-w-mobo-c & 81 & 20180806\_Holy\_sp-s-mobo-m & 73 \\
20180727\_FIRE\_bh-n-mobo-c & 81 & 20180809\_FIRE\_bh-s-mobo-c & 80 \\
20180727\_FIRE\_bh-s-mobo-c & 81 & 20180809\_FIRE\_bl-e-mobo-c & 81 \\
20180727\_FIRE\_bl-e-mobo-c & 81 & 20180809\_FIRE\_mg-w-mobo-c & 81 \\
20180727\_FIRE\_mg-w-mobo-c & 81 & 20180813\_FIRE\_bh-s-mobo-c & 81 \\
20180727\_FIRE\_wc-n-mobo-c & 81 & 20180813\_FIRE\_bl-n-mobo-c & 81 \\
20180728\_FIRE\_rm-w-mobo-c & 81 & 20180813\_FIRE\_mg-w-mobo-c & 81 \\
20180728\_FIRE\_smer-tcs9-mobo-c & 81 & 20180827\_Holyflareup\_sp-e-mobo-c & 81 \\
20180910\_FIRE\_smer-tcs8-mobo-c & 81 & 20180919\_FIRE\_rm-e-mobo-c & 81 \\
20181112\_house\_wc-n-mobo-c & 71 & 20190529\_94Fire\_lp-s-mobo-c & 81 \\
20190529\_94Fire\_om-n-mobo-c & 81 & 20190610\_FIRE\_bh-w-mobo-c & 81 \\
20190610\_Pauma\_bh-w-mobo-c & 80 & 20190610\_Pauma\_bh-w-mobo-m & 80 \\
% Add more rows if necessary
\bottomrule
\end{tabular}
\caption{Scenes composed for target domain dataset by equation \ref{naming_convention} (Part. 4)}
\label{tab:target_domain4}
\end{table}

\begin{table}[htbp]
\centering
\begin{tabular}{llll}
\toprule
\textbf{Scene Description} & \textbf{\# of Imgs} & \textbf{Scene Description} & \textbf{\# of Imgs} \\
\midrule
20190620\_FIRE\_rm-w-mobo-c & 81 & 20190715\_MLOSouth1\_lo-s-mobo-c & 81 \\
20190620\_FIRE\_smer-tcs9-mobo-c & 72 & 20190715\_MLOSouth2\_lo-s-mobo-c & 81 \\
20190629\_FIRE\_hp-n-mobo-c & 57 & 20190715\_MLOSouth3\_lo-s-mobo-c & 81 \\
20190712\_CottonwoodFire\_lp-s-mobo-c & 81 & 20190716\_FIRE\_bl-s-mobo-c & 70 \\
20190712\_FIRE\_om-e-mobo-c & 81 & 20190716\_FIRE\_mg-n-mobo-c & 68 \\
20190712\_RockHouse\_wc-e-mobo-c & 79 & 20190716\_FIRE\_so-w-mobo-c & 72 \\
20190714\_MLOSouth\_lo-s-mobo-c & 81 & 20190716\_Meadowfire\_hp-n-mobo-c & 70 \\
20190714\_PinosSouth\_pi-s-mobo-c & 81 & 20190716\_Riverfire\_rm-w-mobo-c & 80 \\
20190717\_FIRE\_lp-n-mobo-c & 81 & 20190717\_FIRE\_pi-w-mobo-c & 81 \\
20190728\_Dehesa\_lp-n-mobo & 80 & 20190728\_FIRE\_om-n-mobo-c & 79 \\
20190728\_FIRE\_sp-n-mobo-c & 81 & 20190801\_Caliente\_om-w-mobo & 81 \\
20190803\_OtaySouth\_lp-s-mobo & 79 & 20190803\_OtaySouth\_om-s-mobo & 79 \\
% Add more rows if necessary
\bottomrule
\end{tabular}
\caption{Scenes composed for target domain dataset by equation \ref{naming_convention} (Part. 5)}
\label{tab:target_domain5}
\end{table}

\begin{table}[htbp]
\centering
\begin{tabular}{llll}
\toprule
\textbf{Scene Description} & \textbf{\# of Imgs} & \textbf{Scene Description} & \textbf{\# of Imgs} \\
\midrule
20190803\_Sage\_om-n-mobo & 73 & 20190814\_FIRE\_om-e-mobo-c & 79 \\
20190805\_FIRE\_sp-e-mobo-c & 77 & 20190814\_FIRE-pi-s-mobo-c & 80 \\
20190809\_PinosSouth\_pi-s-mobo & 41 & 20190825\_FIRE-smer-tcs8-mobo-c & 80 \\
20190810\_SantaFire\_rm-w-mobo & 81 & 20190825\_FIRE\_sm-w-mobo-c & 75 \\
20190813\_FIRE\_69bravo-e-mobo-c & 81 & 20190826\_FIRE\_pi-s-mobo-c & 80 \\
20190813\_Topanga\_69bravo-n-mobo & 81 & 20190826\_FIRE\_rm-w-mobo-c & 80 \\
20190814\_Border\_lp-s-mobo & 80 & 20190826\_FIRE\_smer-tcs9-mobo-c & 80 \\
20190827\_FIRE\_so-w-mobo-c & 81 & 20190829\_FIRE\_bl-n-mobo-c & 81 \\
20190829\_FIRE\_pi-e-mobo-c & 81 & 20190913\_FIRE\_lp-n-mobo-c & 80 \\
20190829\_FIRE\_rm-w-mobo-c & 81 & 20190915\_FIRE\_rm-n-mobo-c & 78 \\
20190829\_FIRE\_smer-tcs8-mobo-c & 76 & 20190922\_FIRE\_ml-w-mobo-c & 81 \\
20190924\_FIRE\_bl-s-mobo-c & 79 & 20190924\_FIRE\_hp-s-mobo-c & 80 \\
% Add more rows if necessary
\bottomrule
\end{tabular}
\caption{Scenes composed for target domain dataset by equation \ref{naming_convention} (Part. 6)}
\label{tab:target_domain6}
\end{table}

\begin{table}[htbp]
\centering
\begin{tabular}{llll}
\toprule
\textbf{Scene Description} & \textbf{\# of Imgs} & \textbf{Scene Description} & \textbf{\# of Imgs} \\
\midrule
20190924\_FIRE\_lo-w-mobo-c & 79 & 20191001\_FIRE\_om-s-mobo-c & 60 \\
20190924\_FIRE\_lp-n-mobo-c & 72 & 20191001\_FIRE\_rm-w-mobo-c & 81 \\
20190924\_FIRE\_ml-w-mobo-c & 80 & 20191001\_FIRE\_smer-tcs9-mobo-c & 80 \\
20190924\_FIRE\_pi-w-mobo-c & 79 & 20191003\_FIRE\_om-s-mobo-c & 77 \\
20190924\_FIRE\_sm-n-mobo-c & 76 & 20191003\_FIRE\_rm-w-mobo-c & 81 \\
20190924\_FIRE\_wc-e-mobo-c & 72 & 20191003\_FIRE\_smer-tcs9-mobo-c & 77 \\
20190924\_FIRE\_wc-s-mobo-c & 70 & 20191005\_FIRE\_bm-e-mobo-c & 79 \\
20190925\_FIRE\_wc-e-mobo-c & 81 & 20191005\_FIRE\_hp-s-mobo-c & 81 \\
20190925\_FIRE\_wc-s-mobo-c & 81 & 20191005\_FIRE\_vo-n-mobo-c & 77 \\
20190930\_FIRE\_om-s-mobo-c & 80 & 20191005\_FIRE\_wc-e-mobo-c & 79 \\
20191001\_FIRE\_bh-w-mobo-c & 79 & 20191005\_FIRE\_wc-n-mobo-c & 78 \\
20191001\_FIRE\_lp-s-mobo-c & 80 & 20191006\_FIRE\_lo-s-mobo-c & 79 \\
20191001\_FIRE\_om-e-mobo-c & 79 & 20191006\_FIRE\_lo-w-mobo-c & 80 \\
20191006\_FIRE\_lp-e-mobo-c & 72 & 20191006\_FIRE\_lp-n-mobo-c & 73 \\
20191006\_FIRE\_lp-s-mobo-c & 73 & & \\
% Add more rows if necessary
\bottomrule
\end{tabular}
\caption{Scenes composed for target domain dataset by equation \ref{naming_convention} (Part. 7)}
\label{tab:target_domain7}
\end{table}

\begin{table}[htbp]
\centering
\newcolumntype{Y}{>{\centering\arraybackslash}X}
\begin{tabularx}{\textwidth}{>{\hsize=2.1\hsize}Y>{\hsize=0.1\hsize}Y>{\hsize=1.7\hsize}Y>{\hsize=0.1\hsize}Y}
\toprule
\textbf{Scene Description} & \textbf{\small Imgs} & \textbf{Scene Description} & \textbf{\small Imgs} \\
\midrule
20191006\_FIRE\_ml-w-mobo-c & 81 & 20200226\_FIRE\_rm-e-mobo-c & 81 \\
20191006\_FIRE\_om-n-mobo-c & 78 & 20200304\_FIRE\_rm-w-mobo-c & 81 \\
20191006\_FIRE\_om-s-mobo-c & 77 & 20200306\_FIRE\_mlo-n-mobo-c & 81 \\
20191006\_FIRE\_pi-s-mobo-c & 78 & 20200306\_FIRE\_ml-s-mobo-c & 81 \\
20191007\_FIRE\_lp-s-mobo-c & 81 & 20200306\_FIRE\_pi-n-mobo-c & 81 \\
20191007\_FIRE\_om-s-mobo-c & 81 & 20200521\_FIRE\_om-n-mobo-c & 81 \\
20191007\_FIRE\_sm-s-mobo-c & 81 & 20200521\_FIRE\_om-s-mobo-c & 81 \\
 20191030\_CopperCanyon\_om-s-mobo-c & 81 & 20200521\_FIRE\_om-w-mobo-c & 81 \\
 20191030\_CopperCanyon\_om-s-mobo-m & 81 &  20200521\_VEGMGMT\_bm-s-mobo-c & 81 \\
20200202\_FIRE\_hp-w-mobo-c & 81 & {\small 20200521\_VEGMGMT\_ml-w-mobo-c} & 81 \\
20200205\_FIRE\_hp-w-mobo-c & 81 & {\small 20200521\_VEGMGMT\_wc-e-mobo-c} & 81 \\
20200206\_FIRE\_ml-s-mobo-c & 81 & 20200529\_StructFire\_wc-e-mobo-c & 80 \\
 {\small 20200601\_WILDLAND-DRILLS\_mlo-e-mobo-c} & 81 & 20200608\_FIRE\_rm-w-mobo-c & 81 \\
 {\small 20200601\_WILDLAND-DRILLS\_mlo-s-mobo-c} & 81 & 20200611\_skyline\_lp-n-mobo-c & 81 \\
20200601\_WILDLAND-DRILLS\_ml-s-mobo-c & 81 & {\footnotesize 20200614\_DrumCanyon\_syp-w-mobo-c} & 81 \\
20200601\_WILDLAND-DRILLS\_om-e-mobo-c & 81 & 20200615\_Rainbow\_rm-e-mobo-c & 81 \\
% Add more rows if necessary
\bottomrule
\end{tabularx}
\caption{Scenes composed for target domain dataset by equation \ref{naming_convention} (Part. 8)}
\label{tab:target_domain8}
\end{table}

\begin{table}[htbp]
\centering
\newcolumntype{Y}{>{\centering\arraybackslash}X}
\begin{tabularx}{\textwidth}{>{\hsize=2.15\hsize}Y>{\hsize=0.05\hsize}Y>{\hsize=1.75\hsize}Y>{\hsize=0.05\hsize}Y}
\toprule
\textbf{Scene Description} & \textbf{\small Imgs} & \textbf{Scene Description} & \textbf{\small Imgs} \\
\midrule
20200618\_FIRE\_om-w-mobo-c & 81 & 20200727\_Border11Fire\_om-e-mobo-c & 75 \\
20200705\_FIRE\_bm-w-mobo-c & 81 & 20200727\_Border11Fire\_om-e-mobo-m & 75 \\
20200705\_FIRE\_wc-n-mobo-c & 81 & {\small 20200806\_BorderFire\_lp-s-mobo-c} & 81 \\
20200709\_Tripp\_hp-n-mobo-c & 81 & 20200806\_BorderFire\_om-e-mobo-c & 81 \\
{\footnotesize 20200712\_USSBonhommeRichard\_sm-w-mobo-c} & 81 & 20200806\_SpringsFire\_lp-w-mobo-c & 62 \\
20200727\_Border11Fire\_lp-s-mobo-c & 75 & 20200806\_SpringsFire\_lp-w-mobo-m & 62 \\
{\small 20200807\_AppleFire-backfire-operation\_hp-n-mobo-c} & 81 & 20200806\_SpringsFire\_om-n-mobo-c & 65 \\
20200808\_OliveFire\_wc-e-mobo-c & 74 & 20200806\_SpringsFire\_om-n-mobo-m & 62 \\
20200812\_LakeFire\_dwpgm-n-mobo-c & 81 & 20200806\_SpringsFire\_sm-e-mobo-c & 65 \\
20200813\_Ranch2Fire\_marconi-n-mobo-c & 73 & 20200813\_SkylineFire\_sp-n-mobo-c & 75 \\
20200813\_Ranch2Fire\_sjh-n-mobo-c & 78 & 20200813\_VictoriaFire\_lp-n-mobo-c & 70 \\
20200813\_Ranch2Fire\_wilson-e-mobo-c & 77 & 20200822\_BrattonFire\_lp-e-mobo-c & 81 \\
20200822\_BrattonFire\_lp-s-mobo-c & 81 & 20200828\_BorderFire\_om-w-mobo-c & 80 \\
{\small 20200822\_SloaneFire\_lp-n-mobo-c} & 81 & 20200828\_BorderFire\_sm-s-mobo-c & 81 \\
20200823\_OakFire\_pi-e-mobo-c & 81 & {\small 20200829\_inside-Mexico\_cp-s-mobo-c} & 81 \\
{\small 20200829\_inside-Mexico\_mlo-s-mobo-c} & 81 & 20200831\_FIRE\_wc-n-mobo-c & 180 \\
{\small 20200905\_ValleyFire\_cp-s-mobo-c} & 0 & {\small 20200905\_ValleyFire\_lp-n-mobo-c} & 73 \\
{\small 20200905\_ValleyFire\_pi-w-mobo-c} & 75 & {\small 20200905\_ValleyFire\_sm-e-mobo-c} & 71 \\
% Add more rows if necessary
\bottomrule
\end{tabularx}
\caption{Scenes composed for target domain dataset by equation \ref{naming_convention} (Part. 9)}
\label{tab:target_domain9}
\end{table}

\begin{table}[htbp]
\centering
\newcolumntype{Y}{>{\centering\arraybackslash}X}
\begin{tabularx}{\textwidth}{>{\hsize=1.95\hsize}Y>{\hsize=0.05\hsize}Y>{\hsize=1.95\hsize}Y>{\hsize=0.05\hsize}Y}
\toprule
\textbf{Scene Description} & \textbf{\small Imgs} & \textbf{Scene Description} & \textbf{\small Imgs} \\
\midrule
20200911\_FIRE\_lp-e-mobo-c & 81 & 20200930\_inMexico\_lp-s-mobo-c & 81 \\
20200911\_FIRE\_mlo-s-mobo-c & 81 & 20200930\_inMexico\_om-e-mobo-c & 81 \\
20200911\_FIRE\_pi-s-mobo-c & 81 & 20201003\_structurefire\_bh-w-mobo-c & 80 \\
{\footnotesize 20200930\_BoundaryFire\_wc-e-mobo-c} & 81 & 20201003\_structurefire\_bm-w-mobo-c & 74 \\
20200930\_DeLuzFire\_rm-w-mobo-c & 61 & 20201013\_FIRE\_cp-s-mobo-c & 79 \\
20201105\_Roundfire\_lp-s-mobo-c & 80 & {\small 20201105\_Roundfire\_om-e-mobo-c} & 81 \\
20201105\_Roundfire\_pi-s-mobo-c & 81 & 20201127\_Hawkfire\_pi-w-mobo-c & 81 \\
20201202\_BondFire-nightime\_sp-w-mobo-c & 75 & 20201205\_typical-range-fire\_sclm-e-mobo-c & 81 \\
{\footnotesize 20201202\_WillowFire-nightime-near-CDF-HQ\_lp-w-mobo-c} & 73 & 20201206\_JEEP-ON-FIRE\_om-w-mobo-c & 70 \\
{\footnotesize 20201202\_WillowFire-nightime-near-CDF-HQ\_om-n-mobo-c} & 77 & 20201207\_La\_bh-s-mobo-c & 81 \\
{\footnotesize 20201202\_WillowFire-nightime-near-CDF-HQ\_sm-n-mobo-c} & 77 & 20201208\_FIRE\_om-s-mobo-c & 80 \\
20201216\_ChaparralFire\_lp-w-mobo-c & 81 & 20201216\_ChaparralFire\_om-n-mobo-c & 81 \\
20201216\_ChaparralFire\_pi-w-mobo-c & 81 & & \\
% Add more rows if necessary
\bottomrule
\end{tabularx}
\caption{Scenes composed for target domain dataset by equation \ref{naming_convention} (Part. 10)}
\label{tab:target_domain10}
\end{table}

\begin{table}[htbp]
\centering
\newcolumntype{Y}{>{\centering\arraybackslash}X}
\begin{tabularx}{\textwidth}{>{\hsize=1.5\hsize}Y>{\hsize=0.5\hsize}Y>{\hsize=1.5\hsize}Y>{\hsize=0.5\hsize}Y}
\toprule
\textbf{Scene Description} & \textbf{\small\# of Imgs} & \textbf{Scene Description} & \textbf{\small\# of Imgs} \\
\midrule
20201216\_ChaparralFire\_sm-n-mobo-c & 81 & 20210113\_Borderfire\_mlo-s-mobo-c & 81 \\
{\small 20201223\_Creekfire\_bh-w-mobo-c} & 81 & {\small 20210113\_Borderfire\_pi-s-mobo-c} & 81 \\
{\small 20210107\_Miguelfire\_om-w-mobo-c} & 80 & {\small 20210115\_Bonitafire\_tp-w-mobo-c} & 71 \\
20210110\_Borderfire\_lp-s-mobo-c & 80 & 20210204\_FIRE\_tp-s-mobo-c & 81 \\
20210209\_FIRE\_hp-e-mobo-c & 78 & 20210302\_FIRE\_lp-e-mobo-c & 81 \\
20210209\_FIRE\_tp-w-mobo-c & 77 & 20210302\_FIRE\_lp-e-mobo-m & 81 \\
20210319\_FIRE\_om-n-mobo-c & 81 & 20210711\_FIRE\_wc-e-mobo-c & 81 \\
20200906-BobcatFire-wilson-e-mobo-c & 82 & 20210810-Lyonsfire-housefire-lp-n-mobo-c & 64 \\
20220210-EmeraldFire-marconi-w-mobo-c & 82 & 20220210-EmeraldFire-signal-s-mobo-c & 82 \\
20220210-EmeraldFire-stgo-w-mobo-c & 82 & 20220214-PrescribedFire-pi-n-mobo-c & 82 \\
20220302-Jimfire-0921-stgo-e-mobo-c & 81 & 20220302-Jimfire-0921-stgo-s-mobo-c & 81 \\
20220302-Jimfire-1101-stgo-e-mobo-c & 81 & 20220405-fire-in-Fallbrook-rm-s-mobo-c & 81 \\
20220405-fire-in-Fallbrook-rm-s-mobo-m & 82 & 20220622-HighlandFire-wc-n-mobo-c & 81 \\
20220622-HighlandFire-wc-n-mobo-m & 82 & 20220713-Lonestarfire-om-w-mobo-c & 72 \\
{\small 20220727-Casnerfire-bm-s-mobo-c} & 82 & {\small 20220831-Border32fire-pi-s-mobo-c} & 65 \\
20220831-Border32fire-pi-s-mobo-m & 66 & 20220905-FairviewFire-bh-n-mobo-c & 81 \\
20220905-FairviewFire-smer-tcs3-mobo-c & 82 & 20220905-FairviewFire-stgo-e-mobo-c & 81 \\
20220905-FairviewFire-tp-w-mobo-c & 81 & 20221116-Willowfire-om-n-mobo-c & 81 \\
{\small 20221116-Willowfire-sm-n-mobo-c} & 81 & {\small 20230128-Cardboard-Fire-om-w-mobo-c} & 82 \\
% Add more rows if necessary
\bottomrule
\end{tabularx}
\caption{Scenes composed for target domain dataset by equation \ref{naming_convention} (Part. 11)}
\label{tab:target_domain11}
\end{table}

\begin{table}[htbp]
\centering
% Define new column types with centered text
\newcolumntype{Y}{>{\centering\arraybackslash}X}
\begin{tabularx}{\textwidth}{>{\hsize=1.5\hsize}Y>{\hsize=0.5\hsize}Y>{\hsize=1.5\hsize}Y>{\hsize=0.5\hsize}Y}
\toprule
\textbf{Scene Description} & \textbf{\small\# of Imgs} & \textbf{Scene Description} & \textbf{\small\# of Imgs} \\
\midrule
rm-n-mobo-c & 81 & smer-tcs3-mobo-c & 81 \\
lp-e-iqeye & 41 & om-e-mobo-c & 81 \\
pi-s-mobo-c  & 81 & ml-n-mobo-c  & 81 \\
lp-n-iqeye & 41 & mg-s-iqeye & 41 \\
mw-e-mobo-c & 81 & &  \\
\bottomrule
\end{tabularx}
\caption{Scenes composed for source domain dataset by only cameraName}
\label{source domain scenes and number of images using cameraName}
\end{table}

% Begin of Table
\begin{table}[htbp]
\centering
\begin{tabularx}{\textwidth}{XXXX}
\toprule
\textbf{Scene Description} & \textbf{\# of Imgs} & \textbf{Scene Description} & \textbf{\# of Imgs} \\
\midrule
 &  & bl-n-mobo-c & 243 \\
bm-n-mobo-c & 162 & bm-w-mobo-c & 236 \\
rm-w-mobo-c & 1193 & lp-s-iqeye & 81 \\
om-s-mobo-c & 834 & pi-w-mobo-c & 478 \\
sm-n-mobo-c & 628 & bh-w-mobo-c & 559 \\
hp-n-mobo-c & 694 & bm-n-mobo & 81 \\
% Continue adding rows as needed
\bottomrule
\end{tabularx}
\caption{Scenes composed for target domain dataset by only CameraName (Part 1.)}
\label{tab:target domain scenes and number of images for cameraName1}
\end{table}
% End of Table

% Begin of Table
\begin{table}[htbp]
\centering
\begin{tabularx}{\textwidth}{XXXX}
\toprule
\textbf{Scene Description} & \textbf{\# of Imgs} & \textbf{Scene Description} & \textbf{\# of Imgs} \\
\midrule
syp-n-mobo-c & 81 & syp-n-mobo-m & 80 \\
bl-e-mobo-c & 243 & bl-s-mobo-c & 311 \\
bm-s-mobo-c & 227 & sdsc-e-mobo-c & 162 \\
smer-tcs8-mobo-c & 719 & hp-e-mobo-c & 159 \\
mg-n-iqeye & 81 & so-s-mobo-c & 81 \\
bh-n-mobo-c & 402 & lo-s-mobo-c & 565 \\
tp-s-mobo-c & 162 & smer-tcs9-mobo-c & 795 \\
hp-w-mobo-c & 243 & rm-e-mobo-c & 405 \\
pi-e-mobo-c & 405 & rm-s-mobo & 81 \\
smer-tcs10-mobo-c & 243 & sp-s-mobo-c & 153 \\
sp-w-mobo-c & 156 & sm-w-mobo-c & 237 \\
ml-s-mobo-c & 324 & hp-s-mobo-c & 229 \\
wc-e-mobo-c & 939 & sm-e-mobo-c & 217 \\
lp-n-mobo-c & 756 & syp-w-mobo-c & 243 \\
sp-n-mobo-c & 237 & tp-e-mobo-c & 81 \\
sp-e-mobo-c & 239 & so-n-mobo-c & 81 \\
so-w-mobo-c & 234 & bh-s-mobo-c & 323 \\
mg-w-mobo-c & 243 & wc-n-mobo-c & 572 \\
mg-s-mobo-c & 78 & vo-w-mobo-c & 81 \\
sp-s-mobo-m & 73 & lp-s-mobo-c & 874 \\
om-n-mobo-c & 704 & bh-w-mobo-m & 80 \\
mg-n-mobo-c & 68 & lp-n-mobo & 80 \\
om-w-mobo & 81 & lp-s-mobo & 159 \\
om-s-mobo & 79 & om-n-mobo & 73 \\
pi-s-mobo & 41 & rm-w-mobo & 81 \\
69bravo-e-mobo-c & 81 & 69bravo-n-mobo & 81 \\
ml-w-mobo-c & 323 & lo-w-mobo-c & 159 \\
wc-s-mobo-c & 151 & bm-e-mobo-c & 79 \\
vo-n-mobo-c & 77 & lp-e-mobo-c & 315 \\
sm-s-mobo-c & 162 & om-s-mobo-m & 81 \\
mlo-n-mobo-c & 81 & pi-n-mobo-c & 163 \\
om-w-mobo-c & 546 & mlo-e-mobo-c & 81 \\
mlo-s-mobo-c & 324 & lp-s-mobo-m & 75 \\
om-e-mobo-m & 75 & lp-w-mobo-c & 297 \\
lp-w-mobo-m & 62 & om-n-mobo-m & 62 \\
sm-e-mobo-m & 63 & dwpgm-n-mobo-c & 81 \\
marconi-n-mobo-c & 73 & sjh-n-mobo-c & 78 \\
wilson-e-mobo-c & 159 & cp-s-mobo-c & 160 \\
sclm-e-mobo-c & 81 & tp-w-mobo-c & 229 \\
lp-e-mobo-m & 81 & marconi-w-mobo-c & 82 \\
signal-s-mobo-c & 82 & stgo-w-mobo-c & 82 \\
stgo-e-mobo-c & 243 & stgo-s-mobo-c & 162 \\
rm-s-mobo-c & 81 & rm-s-mobo-m & 82 \\
wc-n-mobo-m & 82 & pi-s-mobo-m & 66 \\
smer-tcs3-mobo-m & 82 &  &  \\
\bottomrule
\end{tabularx}
\caption{Scenes composed for target domain dataset by only CameraName (Part 2.)}
\label{tab:target domain scenes and number of images for cameraName2}
\end{table}

\begin{figure}[ht]
    \centering
    \begin{subfigure}{}
        \centering
        \includegraphics[width=0.45\linewidth]{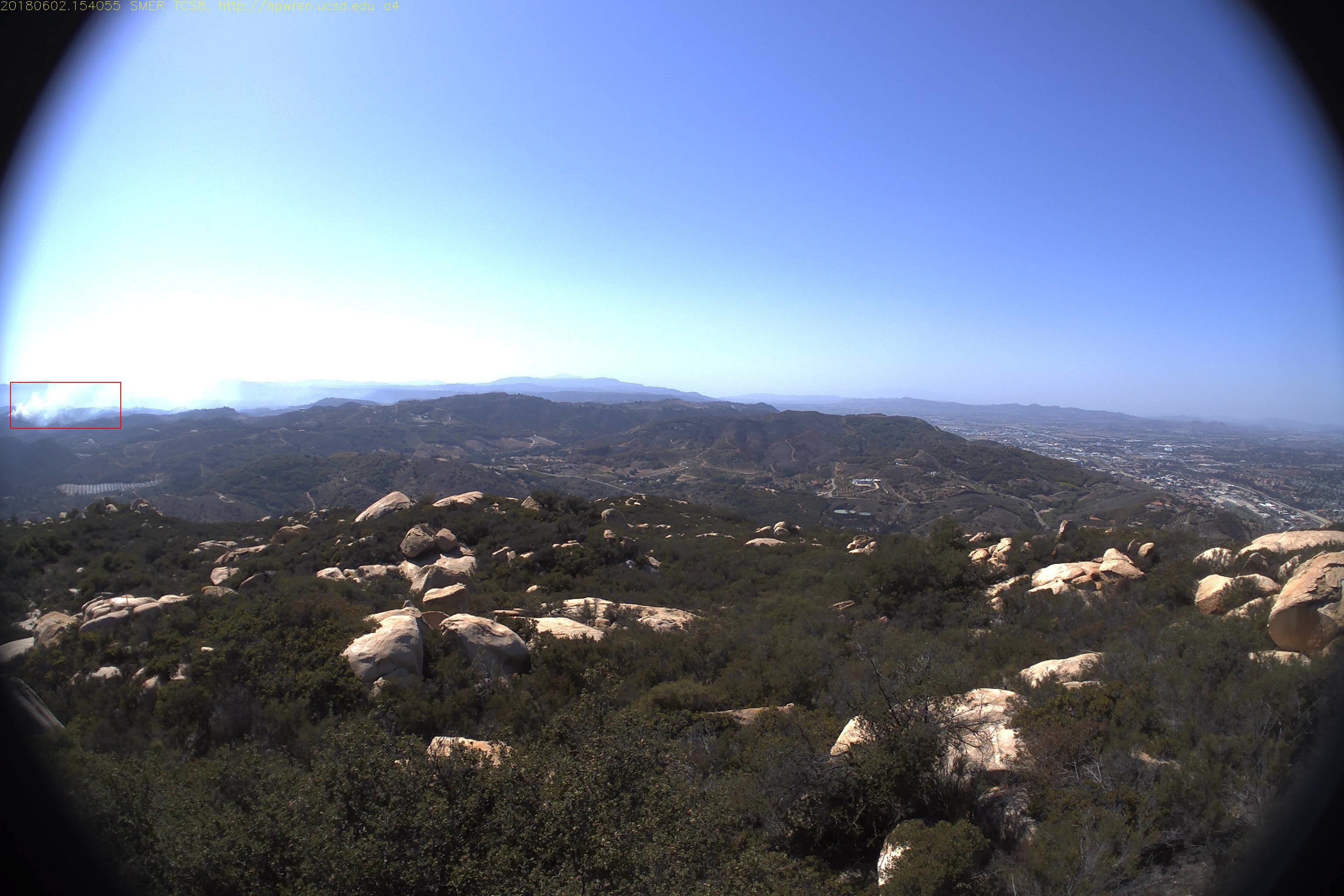}
        \label{fig:subfig_a}
    \end{subfigure}
    \begin{subfigure}{}
        \centering
        \includegraphics[width=0.45\linewidth]{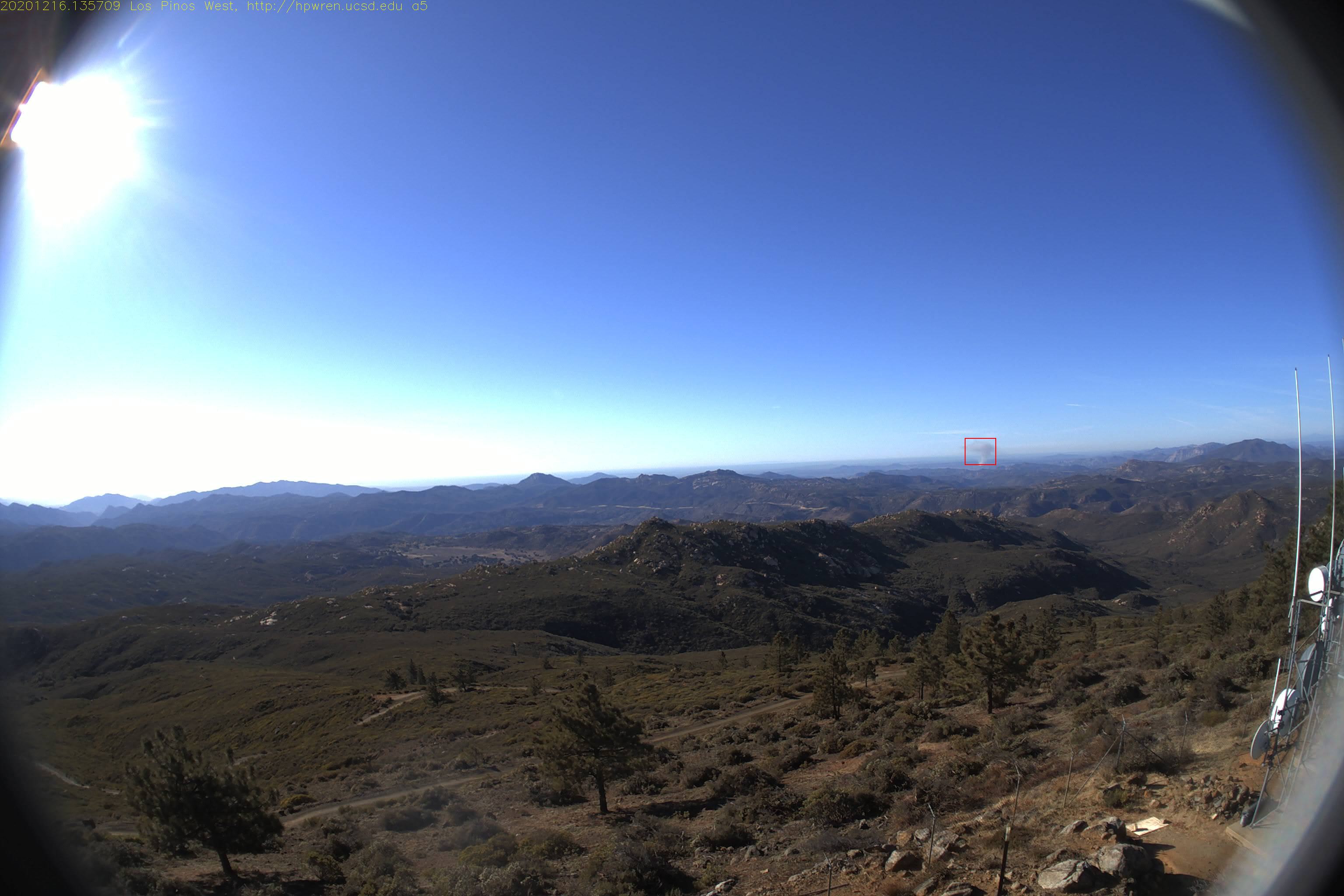}
        \label{fig:subfig_b}
    \end{subfigure}
    \begin{subfigure}{}
        \centering
        \includegraphics[width=0.45\linewidth]{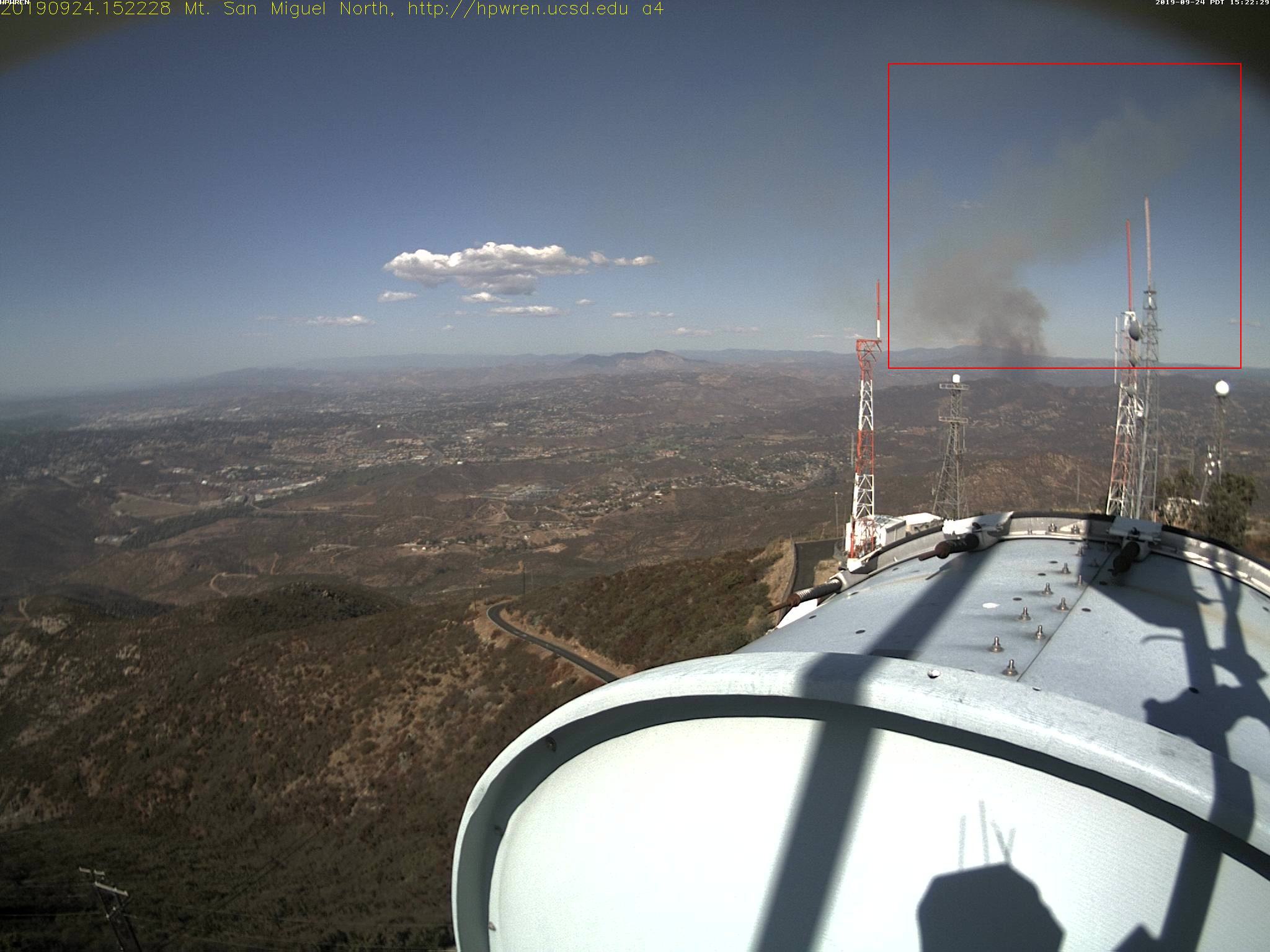}
        \label{fig:subfig_c}
    \end{subfigure}
    \begin{subfigure}{}
        \centering
        \includegraphics[width=0.45\linewidth]{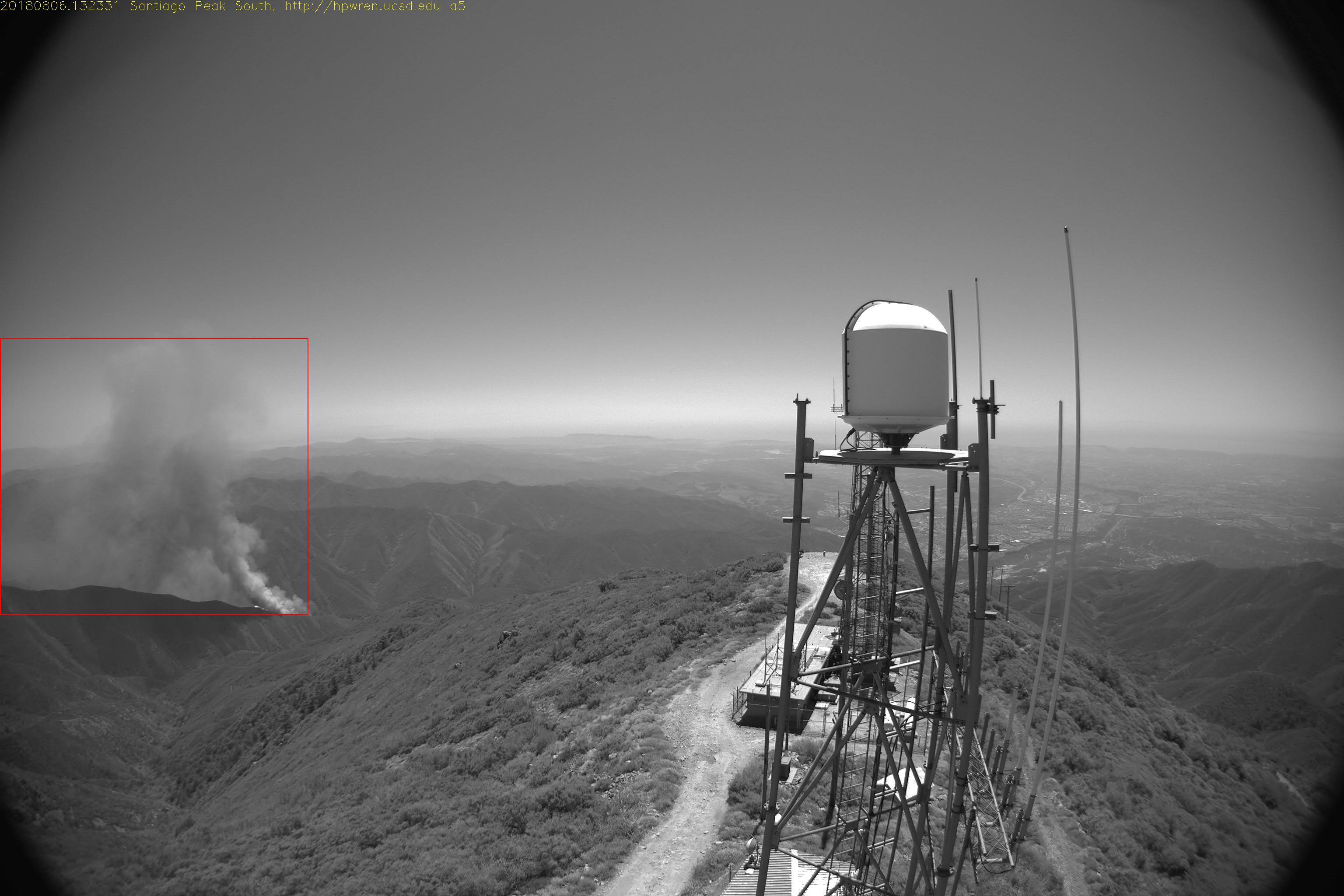}
        \label{fig:subfig_d}
    \end{subfigure}
    \caption{Example of our labeled dataset\centering}
    \label{fig:wildfire_imgs}
\end{figure}

An example of our proposed labels are show in Fig. \ref{fig:wildfire_imgs}.

\section{Training details}
\label{app:Training details}
% Masked ratio, epoch, Domain Adaptation rate, optimizer, 2 step of training

In equation \ref{total_loss}, $L^S, L^M, L^A, L^C$ refers to supervised loss, masked image consistency loss (MIC loss), adversarial loss, and consistency loss each. More information could be seen in  baseline method \cite{Hoyer2022MIC} since we use same losses.

In the first source-only stage, we only used supervised learning. Our backbone is Resnet-50 with ImageNet pretrained weight. We trained 10 epoch with step decay learning rate schedule. We used 16 batch size, SGD optimizer with 0.9 momentum and 0.0005 weight decay. \\
In second stage, we also trained 10 epoch with same hyperparameter setup as first stage except that we used semi-supervised learning loss. We splitted image into 32x32 blocks and masked with ratio 0.5 for strong augmentation. We used 0.9 as EMA rate, and $\lambda^M$ for 0.5. We used confidence score threshold as $\tau_u=0.8, \tau_l=0.05$. The final weight of first stage is used as initial weights. Overall hyperparameters used are summarized in Table \ref{tab:hyp}.

\begin{equation}
min_{\theta_s} \frac{1}{N_s}\sum_{k=1}^{N_s} L_k^S + \frac{1}{N_t}\sum_{k=1}^{N_t} (\lambda^M L_k^M)+\frac{1}{N_t+N_s}\sum_{k=1}^{N_t+N_s}(\lambda^A L_k^A+\lambda^C L_k^C)
\label{total_loss}
\end{equation}

\begin{table}[ht]
\centering
\caption{Hyperparameters of training LADA with the proposed SSDA}
\begin{tabular}{lc}
\hline
\textbf{Config}            &   \\
\hline
Optimizer                  & SGD            \\
Optimizer momentum         & 0.9 \\
Weight decay               & 1e-4             \\
Domain Adaptation rate              & 2.5e-3             \\
Warmup epochs              & 0.333                \\
Training epochs            & 10               \\
EMA decay                  & 0.9            \\
$\tau_u$       & 0.8     \\
$\tau_l$       & 0.05     \\
$\lambda^M$                 & 0.5     \\
$\lambda^A_{ins}$           & 1e-1     \\
$\lambda^A_{img}$           & 2.5e-2     \\
$\lambda^C_{ins}$           & 1e-2     \\
$\lambda^C_{img}$           & 2.5e-3   \\  
\hline
\label{tab:hyp}
\end{tabular}
\end{table}

\section{Impact of using background images for semi-supervised Domain Adaptation}
\label{app:background}
Background images are especially helpful for 0.5\%, 1.0\% protocols where pseudo labels with high confidence score is especially lacking as show in Table \ref{tab:background_image}. We defined background image when there are no object with confidence score higher than 0.05.

\begin{table}[h]
  \centering
  \caption{LADA vs. baseline for ssda protocol}
  \begin{tabular}{|c|c|c|c|}
    \hline
      & \textbf{ssda 0.5\%} & \textbf{ssda 1.0\%} & \textbf{ssda 3.0\%} \\
    \hline
    \textbf{LADA(ours)} & \textbf{10.0} & \textbf{14.0} & 20.4 \\
    \hline
    \textbf{LADA(without background images)} & 7.9 & 13.9 & \textbf{20.9} \\
    \hline
    
  \end{tabular}
  \label{tab:background_image}
\end{table}

\section{Relation between foreground-image and background-image ratio}
\label{app:unlabel_ratio}
% label & unlabel ratio
Since there are dataset imbalance between label and unlabel images, we studied the best ratio between labeled and unlabeled data composing in a mini-batch. As shown in Table \ref{tab:bl_ratio}, 80\% of unlabeled images used for minibatch had best result. It is as expected since the number of unlabeled images are more than 10 times than that of labeled dataset. We found that 90\% unlabeled images used for mini-batch did not converged. We assume it is due to lack of supervisory signal in the early phase. We reported our result based on 80\% of unlabel ratio in a mini-batch.

% table 6
\begin{table}[h]
  \centering
  \caption{ratio between label vs. unlabel iamge in a minibatch}
  \begin{tabular}{|c|c|c|c|c|}
    \hline
      & \textbf{0.5} & \textbf{0.6} & \textbf{0.7} & \textbf{0.8} \\
    \hline
    \textbf{SSDA-0.5\%} & 25.2 & 27.2 & 27.0 & \textbf{27.3} \\    
    \hline

  \end{tabular}
  \label{tab:bl_ratio}
\end{table}
\end{appendix}

\end{document}